\documentclass[sigconf]{acmart}

\usepackage{caption}
\usepackage{algorithm}
\usepackage{algorithmic}

\usepackage{newfloat}
\usepackage{listings}

\usepackage{bm}
\usepackage{multirow}
\usepackage{arydshln}
\usepackage{rotating}
\usepackage{diagbox}
\usepackage{makecell}
\usepackage{colortbl}

\usepackage{enumitem}

\usepackage{xspace}

\newcommand{\model}{\emph{GNNMoE}\xspace}

\AtBeginDocument{%
  }

\copyrightyear{2025}
\acmYear{2025}
\setcopyright{acmlicensed}
\acmConference[WWW Companion '25]{Companion Proceedings of the ACM Web Conference 2025}{April 28-May 2, 2025}{Sydney, NSW, Australia}
\acmBooktitle{Companion Proceedings of the ACM Web Conference 2025 (WWW Companion '25), April 28-May 2, 2025, Sydney, NSW, Australia}
\acmDOI{10.1145/3701716.3715462}
\acmISBN{979-8-4007-1331-6/25/04}

\begin{document}

\title{Mixture of Experts Meets Decoupled Message Passing: Towards General and Adaptive Node Classification}



\author{Xuanze Chen}
\authornote{Both authors contributed equally to this research.}
\author{Jiajun Zhou}
\authornotemark[1]
\authornote{Corresponding Author.}
\affiliation{%
  \institution{Zhejiang University of Technology}
  \city{Hangzhou}
  \country{China}}
  \email{{chenxuanze, jjzhou}@zjut.edu.cn}



\author{Shanqing Yu}
\affiliation{%
  \institution{Zhejiang University of Technology}
  \city{Hangzhou}
  \country{China}}
\email{yushanqing@zjut.edu.cn}

\author{Qi Xuan}
\affiliation{%
  \institution{Zhejiang University of Technology}
  \city{Hangzhou}
  \country{China}}
\email{xuanqi@zjut.edu.cn}

\renewcommand{\shortauthors}{Xuanze Chen, Jiajun Zhou, Shanqing Yu, and Qi Xuan}

\begin{abstract}
  Graph neural networks excel at graph representation learning but struggle with heterophilous data and long-range dependencies. And graph transformers address these issues through self-attention, yet face scalability and noise challenges on large-scale graphs. To overcome these limitations, we propose \model\footnote[1]{Code is available at \url{https://github.com/GISec-Team/GNNMoE}.}, a universal model architecture for node classification. This architecture flexibly combines fine-grained message-passing operations with a mixture-of-experts mechanism to build feature encoding blocks. Furthermore, by incorporating soft and hard gating layers to assign the most suitable expert networks to each node, we enhance the model's expressive power and adaptability to different graph types. In addition, we introduce adaptive residual connections and an enhanced FFN module into \model, further improving the expressiveness of node representation. Extensive experimental results demonstrate that \model performs exceptionally well across various types of graph data, effectively alleviating the over-smoothing issue and global noise, enhancing model robustness and adaptability, while also ensuring computational efficiency on large-scale graphs.
\end{abstract}




\maketitle

\section{Introduction}

Graph Neural Networks (GNNs) have emerged as a powerful tool for representation learning on graph-structured data, leveraging message-passing mechanisms to capture intricate structural information. Despite their success, traditional GNNs face notable challenges, particularly in heterophilous graphs where their underlying assumption of homophily often leads to suboptimal performance~\cite{pathmlp}. Moreover, GNNs are prone to the over-smoothing problem~\cite{PT} when handling long-range dependencies, causing node representations to become indistinguishable after multiple layers of propagation, which reduces their expressiveness.

To address these limitations, Graph Transformer (GT) has gained traction as a promising alternative. By adopting self-attention mechanisms, GT efficiently captures long-range dependencies in graphs, alleviating the over-smoothing issues inherent in traditional GNNs. Moreover, it demonstrates robustness in dealing with both structural and attribute heterogeneity, showing notable performance improvements in various tasks. 
However, despite its excellent performance in certain tasks, GT still faces several critical issues, primarily in two aspects. 
First, the self-attention mechanism, while powerful, may inadvertently introduce significant irrelevant global noise. This issue becomes pronounced in complex graph structures or graphs with weak node relationships, undermining the model's overall effectiveness. Second, the computational complexity of self-attention scales quadratically with graph size, posing severe bottlenecks for processing large-scale graphs and limiting GT's scalability in real-world applications.
For the former issue, several studies~\cite{min2022transformer,dwivedi2021generalization} incorporate topological information as masks or biases to recalibrate self-attention distributions, effectively reducing noise to some extent. For the latter, several studies~\cite{SGFormer,ASN-GT} either optimize the complexity of self-attention computation or compress the graph size to improve scalability. However, despite the observed performance improvement, these studies persist in applying the GT architecture to node classification tasks without thoroughly investigating whether the current GT architecture is genuinely adaptability for this purpose.

Currently, researchers often rely on prior knowledge of graph type to design specific graph representation learning models, requiring extensive experimentation to fine-tune architectures and parameters. This practice, while effective, lacks flexibility and makes it challenging to generalize across diverse graph types efficiently. Therefore, the development of a universal, adaptive, and computationally efficient graph model architecture is a pressing need. Such a model should excel in both homophilous and heterophilous settings, address over-smoothing, minimize global noise from self-attention, and maintain scalability for large-scale graphs.

\begin{figure*}[!htb]
  \centering
  \includegraphics[width=0.8\textwidth]{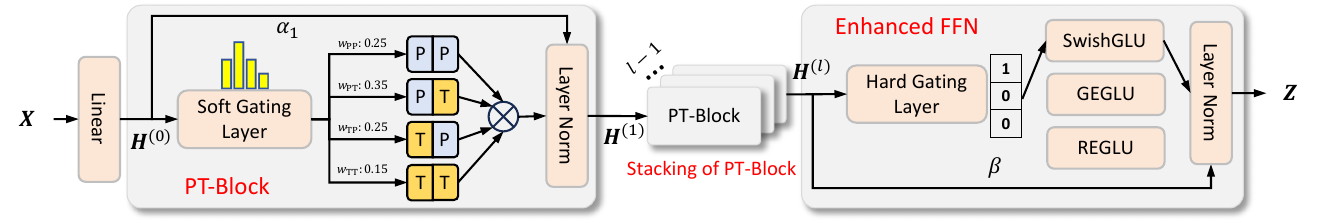}
  \caption{Illustration of \model architectures.}
  \label{fig: architecture}
\end{figure*}
Building upon this motivation, we propose \textbf{\model}, a universal node classification model designed to overcome these challenges. Our architecture decouples the message-passing process, allowing for flexible combinations of fine-grained operations tailored to different graph types. Inspired by the mixture of experts (MoE)~\cite{moe}, \model introduces an adaptive architecture search mechanism that assigns the most suitable expert combinations to each node, thereby enhancing expressiveness and adaptability. Additionally, drawing from residual connections and feed-forward network (FFN) designs in Transformer architectures, we incorporate adaptive residual connections and enhanced FFN modules into \model. These additions further refine the node representation process, boosting classification performance.
Extensive experiments demonstrate that our \model consistently outperforms existing methods across a diverse range of graph datasets. Whether dealing with homophilous or heterophilous graphs, our framework effectively alleviates global noise and over-smoothing issues, showcasing superior robustness and adaptability. Furthermore, through its flexible message-passing operations and adaptive expert combination, \model achieves high performance in node classification tasks while maintaining computational efficiency, even for large-scale graphs.

\section{Preliminaries}
A graph is denoted as $G = (V, E, \boldsymbol{X}, \boldsymbol{Y})$, where $V$ and $E$ are the set of nodes and edges respectively, $\boldsymbol{X} \in \mathbb{R}^{|V| \times d}$ is the node feature matrix, and $\boldsymbol{Y} \in\mathbb{R}^{|V| \times C}$ is the node label matrix. 
Here we use $|V|$, $d$ and $C$ to denote the number of nodes, the dimension of the node features, and the number of classes, respectively. The graph topology information $(V, E)$ can also be denoted by an adjacency matrix $\boldsymbol{A} \in \mathbb{R}^{|V| \times |V|}$, where $\boldsymbol{A}_{ij}=1$ indicates the existence of an edge between $v_i$ and $v_j$, and $\boldsymbol{A}_{ij}=0$ otherwise. 
Node classification is a fundamental task in graph machine learning, and it involves assigning labels to the nodes of a graph based on their features and the graph topology structure.

From a decoupled perspective, message passing in GNNs can be decomposed into two functionally independent operations~\cite{PT}:
\begin{equation}
    \begin{aligned}
         &\textbf{propagation:}\ \  \boldsymbol{h}_\textit{i}^{(l)} = \mathbf{P}\left(\boldsymbol{h}_\textit{i}^{(l-1)},\  \left\{\boldsymbol{h}_\textit{j}^{(l-1)}\mid j\in\mathcal{N}(i) \right\}\right)\\
         &\textbf{transformation:}\ \  \boldsymbol{h}_\textit{i}^{(l)}=\mathbf{T}\left(\boldsymbol{h}_\textit{i}^{(l)}\right)
    \end{aligned}
\end{equation}
where $\boldsymbol{h}_\textit{i}^{(l)}$ is the node representation during $l$-th message passing, $\mathcal{N}(i)$ is the neighbor set of node $v_\textit{i}$. $\mathbf{P}$ is the propagation function that combines message generation and aggregation from neighbor node $v_j$ to target node $v_i$. $\mathbf{T}$ performs a non-linear transformation on the state of the nodes after propagation.
Based on the disentanglement, existing GNN architectures can be loosely categoried into four types according to the stacking order of propagation and transformation operations: $\mathbf{PTPT}$, $\mathbf{PPTT}$, $\mathbf{TTPP}$, and $\mathbf{TPTP}$.

\section{Method: GNNMoE}
To achieve universal node classification across different graphs, we propose \model, as illustrated in Figure~\ref{fig: architecture}. This framework combines the advantages of both GTs and GNNs, consisting of stackable PT-Blocks and enhanced FFN. It takes node features and adjacency information as input and outputs the final node representations. The main highlights are reflected in the following aspects: 1) utilizing a soft gating mechanism to adaptively combine contributions from different message passing experts for each node, enabling flexible processing of different graph messages; 2) employing a hard gating mechanism to select appropriate activation layers, enhancing the expressiveness of FFN; 3) utilizing adaptive residual connections to improve adaptability to different data characteristics.

First, the input features $\boldsymbol{X}$ will be transformed into an initial feature embedding through a linear transformation parameterized by $\boldsymbol{W}_0\in\mathbb{R}^{d\times d^\prime}$ and a ReLU activation:
\begin{equation}
  \boldsymbol{H}^{(0)}=\operatorname{ReLU}\left(\boldsymbol{X}\boldsymbol{W}_0\right)
\end{equation}
where $d^\prime$ is the hidden dimension.
Next, we stack a series of message passing blocks, called PT-blocks, to further learn node representations. Each PT-block consists of a soft gating layer $\operatorname{SG}(\cdot)$, an expert network $\mathcal{E}(\cdot)$, a layer normalization operation $\operatorname{LN}(\cdot)$, and an adaptive initial residual connection, where the expert network $\mathcal{E}=\left\{\text{PP}, \text{PT}, \text{TP}, \text{TT}\right\}$ contains four message passing experts specialized in handling different graph features. For the $(l)$-th PT-block, it takes the node representation output from the $(l-1)$-th PT-block as input, then calculates the allocation weights of the expert network through the soft gating mechanism: 
\begin{equation}
  \boldsymbol{w}_\text{sg}=\operatorname{SG}\left(\boldsymbol{H}^{(l-1)}\right) = \operatorname{Softmax}\left(\boldsymbol{W}_2 \cdot \operatorname{ReLU}\left( \boldsymbol{H}^{(l-1)}\boldsymbol{W}_1\right)\right)
\end{equation}
where $\boldsymbol{w}_\text{sg}\in\mathbb{R}^4$ is the allocation weights, $\boldsymbol{W}_1$ and $\boldsymbol{W}_2$ are the transformation weights.
Next, the graph messages processed by different experts are aggregated using allocation weights, and new node representations are generated through residual connections:
\begin{equation}
  \begin{aligned}
  &\boldsymbol{H}^{(l-1)} = \sum_{i=1}^{4} \boldsymbol{w}_\text{sg}^\textit{i} \cdot \mathcal{E}_\textit{i}\left(\boldsymbol{A}, \boldsymbol{H}^{(l-1)}\right) \\
  &\boldsymbol{H}^{(l)}= \operatorname{LN} \left( \alpha_l \cdot \boldsymbol{H}^{(0)} + (1 - \alpha_l) \cdot \boldsymbol{H}^{(l-1)} \right)
\end{aligned}
\end{equation}
where $\alpha_\textit{l}$ is a learnable parameter that controls the adaptive initial residual connection.

\begin{table*}
  \centering
  \caption{Node classification results: average test accuracy (\%) $\pm$ standard deviation. The best results are highlighted in bold, while the second-best results are underlined. ``Local Rank'' indicates the average performance ranking across homophilous or heterophilous datasets, ``Global Rank'' indicates the average performance ranking across all datasets.}
  \label{tab: main}
  \renewcommand\arraystretch{1.2}
  \resizebox{\textwidth}{!}{
    \begin{tabular}{c|c|ccccccc|ccccccc|c}
      \hline
      \multicolumn{2}{l|}{\diagbox{Method}{Dataset}}                                      & Computers             & Photo                 & \begin{tabular}[c]{@{}c@{}}Coauthor\\CS\end{tabular} & \begin{tabular}[c]{@{}c@{}}Coauthor\\Physics\end{tabular} & Facebook              & ogbn-arixv            & \begin{tabular}[c]{@{}c@{}}Local\\Rank\end{tabular} & Actor                 & \begin{tabular}[c]{@{}c@{}}Chameleon\\-fix\end{tabular} & Squirrel-fix          & Tolokers              & \begin{tabular}[c]{@{}c@{}}Roman\\-empire\end{tabular} & Penn94       & \begin{tabular}[c]{@{}c@{}}Local\\Rank\end{tabular} & \begin{tabular}[c]{@{}c@{}}Global\\Rank\end{tabular}  \\
      \hline
      \multirow{4}{*}{Vanilla}                                                                               & MLP         & 85.01 $\pm$ 0.84               & 92.00 $\pm$ 0.56                 & 94.80 $\pm$ 0.35                   & 96.11 $\pm$ 0.14                                              & 76.86 $\pm$ 0.34          & 53.46 $\pm$ 0.35          & 18.67~                                              & 37.14 $\pm$ 1.06               & 33.31 $\pm$ 2.32                                            & 34.47 $\pm$ 3.09               & 53.18 $\pm$ 6.35              & 65.98 $\pm$ 0.43                                           & 75.18 $\pm$ 0.35                & 17.33    & 18.00                                                 \\
                                                                                                             & GCN         & 91.17 $\pm$ 0.54               & 94.26 $\pm$ 0.59                 & 93.40 $\pm$ 0.45                   & 96.37 $\pm$ 0.20                                              & 93.98 $\pm$ 0.34          & 69.71 $\pm$ 0.18          & 15.33~                                              & 30.65 $\pm$ 1.06               & 41.85 $\pm$ 3.22                                            & 33.89 $\pm$ 2.61               & 70.34 $\pm$ 1.64              & 50.76 $\pm$ 0.46                                           & 80.45 $\pm$ 0.27                & 18.00    & 16.67                                                 \\
                                                                                                             & GAT         & 91.44 $\pm$ 0.43               & 94.42 $\pm$ 0.61                 & 93.20 $\pm$ 0.64                   & 96.28 $\pm$ 0.31                                              & 94.03 $\pm$ 0.36          & 70.03 $\pm$ 0.42          & 14.50~                                              & 30.58 $\pm$ 1.18               & 43.31 $\pm$ 3.42                                            & 36.27 $\pm$ 2.12               & 79.93 $\pm$ 0.77              & 57.34 $\pm$ 1.81                                           & 78.10 $\pm$ 1.28                & 16.67    & 15.58                                                 \\
                                                                                                             & GraphSAGE   & 90.94 $\pm$ 0.56               & 95.41 $\pm$ 0.45                 & 94.17 $\pm$ 0.46                   & 96.69 $\pm$ 0.23                                              & 93.72 $\pm$ 0.35          & 69.15 $\pm$ 0.18          & 14.17~                                              & 37.60 $\pm$ 0.95               & 44.94 $\pm$ 3.67                                            & 36.61 $\pm$ 3.06               & 82.37 $\pm$ 0.64              & 77.77 $\pm$ 0.49                                           & OOM                             & 10.83     & 12.50                                                 \\
      \hline
      \multirow{7}{*}{\begin{tabular}[c]{@{}c@{}}Hetero-\\GNN\end{tabular}}                                  & H2GCN       & 91.69 $\pm$ 0.33               & 95.59 $\pm$ 0.48                 & 95.62 $\pm$ 0.27                   & 97.00 $\pm$ 0.16                                              & 94.36 $\pm$ 0.32          & OOM                       & 7.00~                                               & 37.27 $\pm$ 1.27               & 43.09 $\pm$ 3.85                                            & 40.07 $\pm$ 2.73               & 81.34 $\pm$ 1.16              & 79.47 $\pm$ 0.43                                           & 75.91 $\pm$ 0.44                & 10.83     & 8.92                                                  \\
                                                                                                             & GPRGNN      & 91.80 $\pm$ 0.55               & 95.44 $\pm$ 0.33                 & 95.17 $\pm$ 0.34                   & 96.94 $\pm$ 0.20                                              & 94.84 $\pm$ 0.24          & 69.95 $\pm$ 0.19          & 6.67~                                               & 36.89 $\pm$ 0.83               & 44.27 $\pm$ 5.23                                            & 40.58 $\pm$ 2.00               & 73.84 $\pm$ 1.40              & 67.72 $\pm$ 0.63                                           & 84.34 $\pm$ 0.29                & 9.50     & 8.08                                                  \\
                                                                                                             & FAGCN       & 89.54 $\pm$ 0.75               & 94.44 $\pm$ 0.62                 & 94.93 $\pm$ 0.22                   & 96.91 $\pm$ 0.27                                              & 91.90 $\pm$ 1.95          & 66.87 $\pm$ 1.48          & 15.33~                                              & 37.59 $\pm$ 0.95               & 45.28 $\pm$ 4.33                                            & {\cellcolor[rgb]{1,0.82,0}}\underline{41.05 $\pm$ 2.67}   & 81.38 $\pm$ 1.34              & 75.83 $\pm$ 0.35                                           & 79.01 $\pm$ 1.09                & 8.17     & 11.75                                                \\
                                                                                                             & ACMGCN      & 91.66 $\pm$ 0.78               & 95.42 $\pm$ 0.39                 & 95.47 $\pm$ 0.33                   & 97.00 $\pm$ 0.27                                              & 94.27 $\pm$ 0.33          & 69.98 $\pm$ 0.11          & 7.83~                                               & 36.89 $\pm$ 1.13               & 43.99 $\pm$ 2.02                                            & 36.58 $\pm$ 2.75               & 83.52 $\pm$ 0.87              & 81.57 $\pm$ 0.35                                           & 83.01 $\pm$ 0.46                & 11.00     & 9.42                                                 \\
                                                                                                             & FSGNN       & 91.03 $\pm$ 0.56               & 95.50 $\pm$ 0.41                 & 95.51 $\pm$ 0.32                   & 96.98 $\pm$ 0.20                                              & 94.32 $\pm$ 0.32          & 71.09 $\pm$ 0.21          & 7.33~                                               & 37.14 $\pm$ 1.06               & 45.79 $\pm$ 3.31                                            & 38.25 $\pm$ 2.62               & 83.87 $\pm$ 0.98              & 79.76 $\pm$ 0.41                                           & 83.87 $\pm$ 0.98                & 8.00     & 7.67                                                \\
                                                                                                             & LINKX       & 90.75 $\pm$ 0.36               & 94.58 $\pm$ 0.56                 & 95.52 $\pm$ 0.30                   & 96.93 $\pm$ 0.16                                              & 93.84 $\pm$ 0.32          & 66.16 $\pm$ 0.27          & 12.17~                                              & 31.17 $\pm$ 0.61               & 44.94 $\pm$ 3.08                                            & 38.40 $\pm$ 3.54               & 77.55 $\pm$ 0.80              & 61.36 $\pm$ 0.60                                           & {\cellcolor[rgb]{1,0.82,0}}\underline{84.97 $\pm$ 0.46}    & 12.50    & 12.33                                                \\
      \hline
      \multirow{6}{*}{GT}                                                                                    & Vanilla GT  & 84.41 $\pm$ 0.72               & 91.58 $\pm$ 0.73                 & 94.61 $\pm$ 0.30                   & OOM                                                           & OOM                       & OOM                       & 19.67~                                              & 37.08 $\pm$ 1.08               & 44.27 $\pm$ 3.98                                            & 39.55 $\pm$ 3.10               & 72.24 $\pm$ 1.17              & OOM                                                        & OOM                             & 14.33    & 17.00                                                 \\
                                                                                                             & ANS-GT      & 90.01 $\pm$ 0.38               & 94.51 $\pm$ 0.24                 & 93.93 $\pm$ 0.23                   & 96.28 $\pm$ 0.19                                              & 92.61 $\pm$ 0.16          & OOM                       & 17.50~                                              & {\cellcolor[rgb]{1,0.82,0}}\underline{37.80 $\pm$ 0.95}   & 40.74 $\pm$ 2.26                                            & 36.65 $\pm$ 0.80               & 76.91 $\pm$ 0.85              & 80.36 $\pm$ 0.71                                           & OOM                             & 13.00    & 15.25                                                 \\
                                                                                                             & NAGFormer   & 90.22 $\pm$ 0.42               & 94.95 $\pm$ 0.52                 & 94.96 $\pm$ 0.25                   & 96.43 $\pm$ 0.20                                              & 93.35 $\pm$ 0.28          & 70.25 $\pm$ 0.13          & 13.83~                                              & 36.99 $\pm$ 1.39               & 46.12 $\pm$ 2.25                               & 38.31 $\pm$ 2.43               & 66.73 $\pm$ 1.18              & 75.92 $\pm$ 0.69                                           & 73.98 $\pm$ 0.53                & 13.00    & 13.42                                                 \\
                                                                                                             & SGFormer    & 90.70 $\pm$ 0.59               & 94.46 $\pm$ 0.49                 & 95.21 $\pm$ 0.20                   & 96.87 $\pm$ 0.18                                              & 86.66 $\pm$ 0.54          & 65.84 $\pm$ 0.24          & 15.00~                                              & 36.59 $\pm$ 0.90               & 44.27 $\pm$ 3.68                                            & 38.83 $\pm$ 2.19               & 80.46 $\pm$ 0.91              & 76.41 $\pm$ 0.50                                           & 76.65 $\pm$ 0.49                & 13.17    & 14.08                                                 \\
                                                                                                             & Exphormer   & 91.46 $\pm$ 0.51               & 95.42 $\pm$ 0.26                 & 95.62 $\pm$ 0.29                   & 96.89 $\pm$ 0.20                                              & 93.88 $\pm$ 0.40          & 71.59 $\pm$ 0.24          & 8.17~                                               & 36.83 $\pm$ 1.10               & 42.58 $\pm$ 3.24                                            & 36.19 $\pm$ 3.20               & 82.26 $\pm$ 0.41              & {\cellcolor[rgb]{1,0.6,0}}\textbf{87.55 $\pm$ 1.13}                                  & OOM                             & 13.33    & 10.75                                                  \\
                                                                                                             & Difformer   & 91.52 $\pm$ 0.55               & 95.41 $\pm$ 0.38                 & 95.49 $\pm$ 0.26                   & 96.98 $\pm$ 0.22                                              & 94.23 $\pm$ 0.47          & OOM                       & 9.83~                                               & 36.73 $\pm$ 1.27               & 44.44 $\pm$ 3.20                                            & 40.45 $\pm$ 2.51               & 81.04 $\pm$ 4.16              & 72.52 $\pm$ 0.44                                           & OOM                             & 12.67    & 11.25                                                 \\
      \hline
      \multirow{3}{*}{GNNMoE}                                                                                & GCN-like P  & {\cellcolor[rgb]{1,0.6,0}}\textbf{92.17 $\pm$ 0.50}      & {\cellcolor[rgb]{1,0.6,0}}\textbf{95.81 $\pm$ 0.41}        & {\cellcolor[rgb]{1,0.6,0}}\textbf{95.81 $\pm$ 0.26}          & {\cellcolor[rgb]{1,0.82,0}}\underline{97.03 $\pm$ 0.13}                                  & {\cellcolor[rgb]{1,0.6,0}}\textbf{95.53 $\pm$ 0.35} & {\cellcolor[rgb]{1,0.82,0}}\underline{72.29 $\pm$ 0.16}  & {\cellcolor[rgb]{1,0.6,0}}\textbf{1.33}~                                           & 37.59 $\pm$ 1.36               & {\cellcolor[rgb]{1,0.6,0}}\textbf{47.19 $\pm$ 2.93}                                   & {\cellcolor[rgb]{1,0.6,0}}\textbf{44.02 $\pm$ 2.59}      & {\cellcolor[rgb]{1,0.82,0}}\underline{84.77 $\pm$ 0.93}  & 85.05 $\pm$ 0.55                                           & {\cellcolor[rgb]{1,0.6,0}}\textbf{85.11 $\pm$ 0.39}       & {\cellcolor[rgb]{1,0.6,0}}\textbf{2.33}     & {\cellcolor[rgb]{1,0.6,0}}\textbf{1.83}                                                  \\
                                                                                                             & SAGE-like P & 91.85 $\pm$ 0.39               & 95.46 $\pm$ 0.24                 & 95.68 $\pm$ 0.24                   & 96.81 $\pm$ 0.22                                              & 94.63 $\pm$ 0.36          & 71.94 $\pm$ 0.25          & 5.67~                                               & {\cellcolor[rgb]{1,0.6,0}}\textbf{37.97 $\pm$ 1.01}      & 45.73 $\pm$ 3.19                                            & 39.19 $\pm$ 2.84               & 83.96 $\pm$ 0.75              & 86.00 $\pm$ 0.45                                           & 84.05 $\pm$ 0.37                & {\cellcolor[rgb]{1,0.82,0}}\underline{4.17}  & 4.92                                                  \\
                                                                                                             & GAT-like P  & {\cellcolor[rgb]{1,0.82,0}}\underline{91.98 $\pm$ 0.46}   & {\cellcolor[rgb]{1,0.82,0}}\underline{95.71 $\pm$ 0.37}     & {\cellcolor[rgb]{1,0.82,0}}\underline{95.72 $\pm$ 0.23}       & {\cellcolor[rgb]{1,0.6,0}}\textbf{97.05 $\pm$ 0.19}                                     & {\cellcolor[rgb]{1,0.82,0}}\underline{95.21 $\pm$ 0.25}  & {\cellcolor[rgb]{1,0.6,0}}\textbf{72.45 $\pm$ 0.32} & {\cellcolor[rgb]{1,0.82,0}}\underline{1.67~}                                           & 37.76 $\pm$ 0.98               & 45.56 $\pm$ 3.94                                            & 39.19 $\pm$ 2.38               & {\cellcolor[rgb]{1,0.6,0}}\textbf{85.45 $\pm$ 0.94}     & {\cellcolor[rgb]{1,0.82,0}}\underline{87.29 $\pm$ 0.60}                               & 81.98 $\pm$ 0.47                & 4.33              & {\cellcolor[rgb]{1,0.82,0}}\underline{3.00}                                                  \\
      \hline
      \multicolumn{1}{c|}{\multirow{2}{*}{\begin{tabular}[c]{@{}c@{}}GNNMoE\\(GCN-like P)\end{tabular}}}            & w/o FFN     & 91.29 $\pm$ 0.36 & 95.67 $\pm$ 0.37 & 95.49 $\pm$ 0.25                                         & 96.95 $\pm$ 0.17                                              & 94.98 $\pm$ 0.37                             &71.31 $\pm$ 0.29                                    & 6.17~          & 37.26 $\pm$ 0.91                     & 45.22 $\pm$ 4.21                                            & 39.08 $\pm$ 2.19 & 84.17 $\pm$ 0.70 & 84.53 $\pm$ 0.26                                           & 79.80 $\pm$ 1.21                             & 7.67          & 6.92            \\
      \multicolumn{1}{c|}{}                                                                                         & w/o AIR/AR~ & 91.76 $\pm$ 0.32 & 95.33 $\pm$ 0.49 & 94.31 $\pm$ 0.31                                         & 96.75 $\pm$ 0.27                                              & 94.91 $\pm$ 0.40                             &70.88 $\pm$ 0.28                                    & 9.83~          & 37.35 $\pm$ 0.98                     & {\cellcolor[rgb]{1,0.82,0}}\underline{47.09 $\pm$ 3.39}                                            & 39.15 $\pm$ 2.87 & 84.04 $\pm$ 0.70 & 84.88 $\pm$ 0.84                                           & 80.61 $\pm$ 0.97                             & 5.83          & 7.83            \\
      \hline
  \end{tabular}}
\end{table*}

After message passing via $l$ PT-blocks, \model has effectively fused the attribute information of the nodes with the topological information. Furthermore, inspired by the vanilla GT architecture, where adding FFN can enhance the expressiveness of vanilla GNN, we design an enhanced FFN module in the \model architecture. Specifically, the enhanced FFN module consists of a hard gating layer $\operatorname{HG}(\cdot)$, an expert network $\mathcal{A}(\cdot)$, a layer normalization operation, and an adaptive residual connections, where the expert network $\mathcal{A}=\left\{\text{SwishGLU},\text{GEGLU},\text{REGLU} \right\}$ contains three activation function experts. SwishGLU combines Swish activation with gating mechanisms to promote more effective gradient propagation; GEGLU enhances nonlinear expressiveness through additive activation and gating; REGLU introduces gating on top of ReLU to reduce gradient vanishing and improve computational efficiency.

In the enhanced FFN, the node features encoded by $l$ PT-blocks are first input into a hard gating layer, which selects the appropriate activation function expert for further feature encoding:
\begin{equation}
  j=\operatorname{HG}\left(\boldsymbol{H}^{(l)}\right)=\operatorname{Gumbel\_Softmax}\left(\boldsymbol{H}^{(l)}\right) \in\{1,2,3\}
\end{equation}
Then, the selected expert will further encode $\boldsymbol{H}^{(l)}$ to enhance its expressiveness, followed by an adaptive residual connection to generate the final node representation:
\begin{equation}
  \begin{aligned}
    &\boldsymbol{Z} = \mathcal{A}_\textit{j}\left(\boldsymbol{H}^{(l)}\right) =  \left( \sigma_\textit{j}\left(\boldsymbol{H}^{(l)}\boldsymbol{W}_3\right)\otimes \boldsymbol{H}^{(l)}\boldsymbol{W}_4 \right)\boldsymbol{W}_5\\
    &\boldsymbol{Z} = \operatorname{LN}\left(\beta \cdot \boldsymbol{H}^{(0)}+(1-\beta)\cdot \boldsymbol{Z}\right)
  \end{aligned}
\end{equation}
where $\sigma\in \left\{\text{Swish},\text{GELU},\text{ReLU}\right\}$, $\boldsymbol{W}_3$, $\boldsymbol{W}_4$, $\boldsymbol{W}_5\in\mathbb{R}^{d^\prime\times d^\prime}$ are the transformation weights, $\otimes$ is the element-wise multiplication, $\beta$ is a learnable parameter that controls the adaptive residual connection.

To achieve node classification, we finally use a prediction head $f_\text{pred}$ parameterized by $\boldsymbol{W}_6\in\mathbb{R}^{d^\prime \times C}$ and Softmax activation to obtain the node predictions. During model training, binary cross-entropy classification loss is used as the optimization objective.
\begin{equation}
  \hat{\boldsymbol{Y}}=\operatorname{Softmax}\left( \boldsymbol{Z}\boldsymbol{W}_6 \right), \quad
    \mathcal{L} = -\operatorname{trace}\left(\boldsymbol{Y}_\text{train}^\top \cdot \log \hat{\boldsymbol{Y}}_\text{train} \right)
\end{equation}
where the trace operation $\operatorname{trace}\left( \cdot \right)$ is used to compute the sum of the diagonal elements of the matrix.

\section{Experiments}
\subsection{Experiment Settings}
\subsubsection{Datasets and Baselines}
We conduct extensive experiments on 12 benchmark datasets, which include 
(1) Six homophilous datasets: Computers, Photo, Coauthor CS, Coauthor Physics, Facebook and ogbn-arxiv; and
(2) Six heterophilous datasets: Actor, Squirrel-fix, Chameleon-fix~\cite{platonov2023a}, Tolokers, Roman-empire and Penn94.
All datasets are divided into training, validation, and testing sets in a proportion of 48\%: 32\%: 20\%.
We compare \model with three kinds of baselines, which include (1) Vanilla model: MLP, GCN, GAT, GraphSAGE; (2) Heterophilous GNNs: LINKX, H2GCN, GPRGNN, FAGCN, ACMGCN, FSGNN; and (3) GT models: vanilla GT, ANS-GT, NAGFormer~\cite{chennagphormer}, SGFormer~\cite{SGFormer}, Exphormer~\cite{exphormer} and Difformer~\cite{wu2023difformer}.

\subsubsection{Experimental Settings}
We utilize 10 random seeds to fix the data splits and model initialization, and report the average accuracy and standard deviation over 10 runs.
For all methods, we set the search space of common parameters as follows:
maximum epochs to 500 with 100 patience,
hidden dimension $d^\prime$ to 64,
optimizer to AdamW,
learning rate in \{0.005, 0.01, 0.05, 0.1\},
dropout rate in \{0.1, 0.3, 0.5, 0.7, 0.9\}.
For \model, the number of PT-blocks in \{3,4,5,6\} is searched for ogbn-arxiv while a fixed value of 2 is used for other datasets.
For all baselines, we search the common parameters in the same parameter spaces.

\begin{figure}[!htb]
  \centering
  \includegraphics[width=\linewidth]{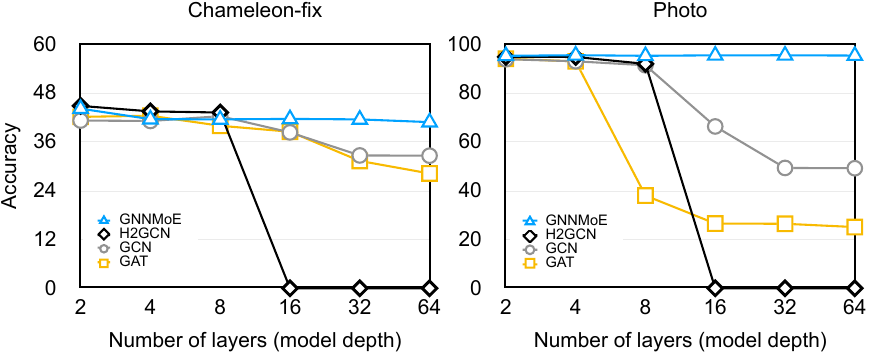}
  \caption{Impact of model depth.}
  \label{fig: over}
\end{figure}

\subsection{Evaluation on Node Classification}
Table~\ref{tab: main} reports the node classification results of all methods, from which we can draw the following conclusions: 1) \model consistently demonstrates higher local and global average performance rankings on both homophilous and heterophilous datasets, indicating its effectiveness, superiority, and stability in node classification tasks, significantly surpassing three categories of baselines; 2) Our method successfully avoids encountering out-of-memory (OOM) issues, in contrast to certain GT-based methods and spatial-domain GNNs, which suggests the architectural efficiency of \model and its scalability in large-scale graph computations.

We further conduct ablation studies to analyze the impact of specific components on the performance of \model. As observed, removing the FFN module (w/o FFN) and the residual connection module (w/o AIR/AR) leads to significant performance degradation across all datasets. This highlights the critical role of the enhanced FFN module in improving \model's utilization of important features, as well as the contribution of the residual connection module in facilitating the model's use of initial information. Both components contribute to \model's universality across different graphs.

\subsection{More Analysis}

\subsubsection{Impact on Model Depth}
Figure~\ref{fig: over} demonstrates the impact of model depth on performance. It is evident that Vanilla GNN's performance rapidly deteriorates as model depth increases, indicating the presence of over-smoothing. Meanwhile, H2GNN's performance gradually declines as model depth increases from 2 to 8 layers, and encounters memory overflow when model depth exceeds 16 layers. In contrast, our method maintains consistently stable performance while stacking PTblock message passing modules, demonstrating its immunity to the over-smoothing problem.

\subsubsection{Efficiency Analysis}
\begin{figure}[!htb]
  \centering
  \includegraphics[width=0.65\linewidth]{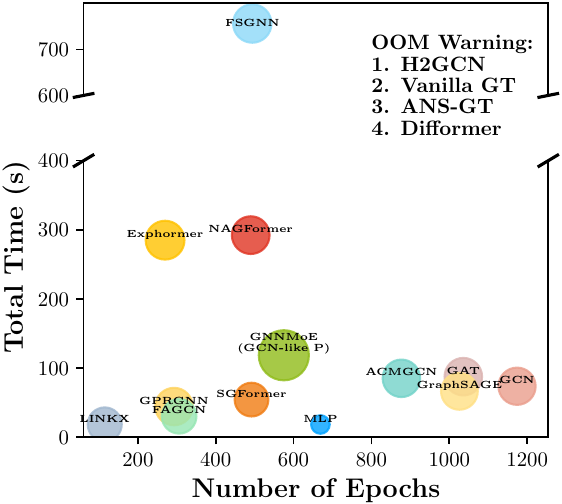}
  \caption{Efficiency analysis on ogbn-arxiv dataset.}
  \label{fig:efficiency}
\end{figure}

Figure~\ref{fig:efficiency} illustrates the efficiency and accuracy of several representative methods on the ogbn-arxiv dataset, where the x-axis represents the number of epochs at which early stopping is triggered, the y-axis represents the total training time, and the bubble size reflects accuracy. As shown, compared to the spatial-domain GNN method FSGNN and some GT-based methods, \model consumes 2-7 times less training time. Additionally, compared to traditional GNN methods, \model converges in fewer epochs. In summary, \model demonstrates good computational efficiency while maintaining high performance.

\section{Conclusion}

We combine the strengths of GNNs and GTs to design a universal node classification model architecture, \model. This architecture encapsulates adaptive message passing into expert network blocks, providing flexible encoding capabilities for different types of graphs. Extensive experiments demonstrate that, compared to existing GNNs, as well as GTs, \model outperforms in node classification performance, while also adapting to various graph types, showcasing its universality. Additionally, our architecture effectively addresses inherent challenges such as over-smoothing and inefficiency. In the future, we will further improve the existing architecture by optimizing both the input and structure of the gating networks, while also exploring graphs in more domains.

\begin{acks}
  This work was supported in part by China Post-Doctoral Science Foundation under Grant 2024M762912, 
  in part by the Post-Doctoral Science Preferential Funding of Zhejiang Province of China under Grant ZJ2024060, 
  in part by the Key R\&D Program of Zhejiang under Grants 2022C01018, 
  in part by the National Natural Science Foundation of China under Grant U21B2001.
\end{acks}

\bibliographystyle{ACM-Reference-Format}
\bibliography{WWW-base}










\end{document}